\PassOptionsToPackage{unicode}{hyperref}
\PassOptionsToPackage{hyphens}{url}
\documentclass[
]{article}
\usepackage{amsmath,amssymb}
\usepackage{lmodern}
\usepackage{iftex}
\usepackage{pifont}
\ifPDFTeX
  \usepackage[T1]{fontenc}
  \usepackage[utf8]{inputenc}
  \usepackage{textcomp} 
\else 
  \usepackage{unicode-math}
  \defaultfontfeatures{Scale=MatchLowercase}
  \defaultfontfeatures[\rmfamily]{Ligatures=TeX,Scale=1}
\fi
\IfFileExists{upquote.sty}{\usepackage{upquote}}{}
\IfFileExists{microtype.sty}{
  \usepackage[]{microtype}
  \UseMicrotypeSet[protrusion]{basicmath} 
}{}
\makeatletter
\@ifundefined{KOMAClassName}{
  \IfFileExists{parskip.sty}{%
    \usepackage{parskip}
  }{
    \setlength{\parindent}{0pt}
    \setlength{\parskip}{6pt plus 2pt minus 1pt}}
}{
  \KOMAoptions{parskip=half}}
\makeatother
\usepackage{xcolor}
\usepackage{graphicx}
\makeatletter
\def\maxwidth{\ifdim\Gin@nat@width>\linewidth\linewidth\else\Gin@nat@width\fi}
\def\maxheight{\ifdim\Gin@nat@height>\textheight\textheight\else\Gin@nat@height\fi}
\makeatother
\setkeys{Gin}{width=\maxwidth,height=\maxheight,keepaspectratio}
\makeatletter
\def\fps@figure{htbp}
\makeatother
\ifLuaTeX
  \usepackage{selnolig}  
\fi
\IfFileExists{bookmark.sty}{\usepackage{bookmark}}{\usepackage{hyperref}}
\IfFileExists{xurl.sty}{\usepackage{xurl}}{} 
\hypersetup{
  hidelinks,
  pdfcreator={LaTeX via pandoc}}

\author{}
\date{}
\usepackage{caption}

\begin{document}

\textbf{Emergence of Symbols in Neural Networks for Semantic Understanding and Communication}

Yang Chen\textsuperscript{1,*,$\dagger$}, Liangxuan Guo\textsuperscript{1,3,$\dagger$}, Shan Yu\textsuperscript{1,2,~*}

\begin{quote}
1 Brainnetome Center and National Laboratory of Pattern Recognition, Institute of Automation, Chinese Academy of Sciences,Beijing, 100190, China.~

2 School of Artificial Intelligence, University of Chinese Academy of Sciences (UCAS), Beijing, 100049, China

3 School of Future Technology, University of Chinese Academy of Sciences (UCAS), Beijing, 100049, China

*Corresponding authors. Email: \href{mailto:yang.chen@ia.ac.cn}{\nolinkurl{yang.chen@ia.ac.cn}} (Y.C.), \href{mailto:shan.yu@nlpr.ia.ac.cn}{\nolinkurl{shan.yu@nlpr.ia.ac.cn}} (S.Y.)

$\dagger$ These authors contributed equally to this work.

\end{quote}

\textbf{Abstract}

The capacity to generate meaningful symbols and effectively employ them for advanced cognitive processes, such as communication, reasoning, and planning, constitutes a fundamental and distinctive aspect of human intelligence. Existing deep neural networks still notably lag human capabilities in terms of generating symbols for higher cognitive functions. Here, we propose a solution (symbol emergence artificial network (SEA-net)) to endow neural networks with the ability to create symbols, understand semantics, and achieve communication. SEA-net generates symbols that dynamically configure the network to perform specific tasks. These symbols capture compositional semantic information that allows the system to acquire new functions purely by symbolic manipulation or communication. In addition, these self-generated symbols exhibit an intrinsic structure resembling that of natural language, suggesting a common framework underlying the generation and understanding of symbols in both human brains and artificial neural networks. We believe that the proposed framework will be instrumental in producing more capable systems that can synergize the strengths of connectionist and symbolic approaches for artificial intelligence (AI).

\textbf{\textsc{MAIN TEXT}}

\textbf{Introduction}

Humans are a symbolic-based species (1). We can proficiently use symbols to understand and communicate about the external world and our internal state, as well as to reason about relationships and plan for actions (2, 3), thus providing humans with a decisive evolutionary advantage. Recently, large language models (LLMs) have demonstrated remarkable progress in sophisticated tasks of natural language processing (4, 5). However, the training of LLMs relies exclusively on the corpus of human language. This reasonable approach may prevent artificial intelligence (AI) systems from developing a language ability as powerful as that of humans. The key missing component in the current approach is the ability to create meaningful symbols through interactions with the environment.
Human language is an embodied symbolic system, meaning that it is intrinsically related to the design of our brain and other parts of our body, as well as the way in which we interact with the physical world (6). An AI agent with a different bodily design that works in a different environment will need to develop its own embodied symbolic system to represent and communicate concepts indescribable in human language.

The symbol emergence problem arises when AI systems attempt to generate symbols through interactions with the environment, similar to the way in which humans create symbols and understand their meaning (7). Specifically,  the problem refers to the ability to create new symbols to 1) facilitate the formation of new concepts from novel experiences (7, 8) and 2) better support embodied cognition (EC) (9–11) or, more broadly, grounded cognition (GC) (6, 12). Importantly, the created symbols must capture compositional semantic information, ensuring that semantic units can be flexibly recombined to express new composite meanings to support knowledge generalization across different contexts (13–15). Simultaneously, the created symbols must be properly grounded, thereby enabling the system to map symbols to specific functions or interval representations, known as the symbol grounding problem (13, 16–18).

In the present study, we propose a new framework (symbol emergence artificial network (SEA-net)) that endows neural networks with the ability to create understandable symbols and acquire new knowledge by symbolic communication among different agents. Interestingly, the semantic structure embedded in the SEA-net-generated system shares important features with the human language, suggesting a potential common mechanism underlying symbol generation and understanding in the human brain and proposed model.

\textbf{Results}

\textbf{Grounding emerging symbols to network structure for semantic understanding}

Here, we propose a novel solution (SEA-net) to solve the symbol emergence and grounding problems for artificial neural networks simultaneously. The SEA-net framework is comprised of two integral components (Fig. 1a): 1) The task solving (TS) module, which can manifest in diverse network structures tailored for specific tasks (e.g., networks for image classification); and 2) The context-dependent processing (CDP) module (19), inspired by the primate prefrontal cortex (PFC) (20) and facilitating flexible processing in response to varying contextual inputs (21).. The CDP serves to link symbols to the network structure in the TS module. Specifically, the CDP module transforms its symbolic inputs into controlling signals to configure the proper state of the TS module, via “gating” or “gain control” operation, to perform specific tasks. Thus, this design grounds the meaning of symbols to specific network configurations (Fig. 1b).

We next  addressed how to generate symbols with meaning grounded . Experiments were conducted based on an identification task (13) using the CIFAR 100 dataset. The task was to decide if an image presented to the TS module was a member of a particular category based on symbolic inputs to the CDP module. The training process consisted of network and symbol training phases (Fig.1b and Fig. 1c). First, a random 20-element vector was used as the symbol assigned to each category. In the network training phase, symbolic inputs to the CDP module were fixed, while all network parameters, including those in both modules, were updated by gradient back-propagation using a binary supervising signal to indicate whether the image belonged to the target category or not. In the symbol training phase, only symbolic inputs were modified by the back-propagated gradients, with all network parameters fixed. These two phases were executed in a round-robin fashion until identification accuracy plateaued. We found that for all 100 categories tested, symbols corresponding to individual categories emerged to achieve a classification accuracy ranging from 0.69 to 0.92, well above the chance level of 0.5 (Fig. 1d). These results indicated that highly compressed symbols (20 elements) could be self-generated and used to configure a complex network (approximately 104 connection weights) to perform specific tasks, suggesting a promising framework for artificial neural networks to create meaningful symbols and dynamically execute various functions according to symbolic inputs.

To examine the internal structure of these emerged symbols, we performed hierarchical clustering based on cosine distance. This analysis revealed a modular structure characterized by distinct symbol clusters (Fig. 2a). Visualization of the modular structure highlighted the relationships between individual categories (Fig. 2b), similar to the semantic network used to study natural language (22). Notably, semantically close categories formed clusters, e.g., clusters of people, animals, trees, fruit, furniture, and automobiles. These symbols enabled the establishment of connections among concepts through diverse multidimensional relationships, including similarities in foreground shape (e.g., snakes and worms), foreground color (e.g., sweet pepper and sunflower), background (e.g., mushrooms and snail), and co-occurrence (e.g., palm tree, cloud, and sea; tulip and butterfly), etc.

\textbf{Learning new tasks by symbolic manipulation}

Next, we examined whether the symbols generated by SEA-net, along with their corresponding internal network configurations, could effectively capture compositional semantic information that enables the network to learn new capabilities by searching the symbol space alone, without changing the network structure. To this end, we conducted an incremental learning experiment. Specifically, in the first stage, the system was trained to recognize 99 classes through training network parameters and symbolic inputs together, as described above. In the second stage, the remaining class (referred to as the symbolically inferred class) was learned by manipulating the symbolic inputs to the system, while the network parameters remained fixed. Results showed that all categories were symbolically inferred with accuracy well beyond chance (Fig. S1). Thus, the system acquired new capabilities by pure symbolic manipulation, i.e., SEA-net can form novel symbols, based on a grounded symbol set, to correctly identify new categories.

Interestingly, certain categories (e.g., bee, flatfish, keyboard, and worm) were identified with significantly higher accuracy when the network parameters and symbolic inputs were trained simultaneously, rather than by symbolic manipulation alone, indicating that these categories cannot be effectively inferred by leveraging knowledge obtained from learning other categories. Consistently, these categories were more often located at the periphery of the semantic network, i.e., at a further distance from other nodes (Fig. S2). Previous research has suggested that a specific set of symbols form the “grounding kernel” (23), thus creating a highly parsimonious foundation to composite other symbols. Those categories that cannot be well inferred symbolically may belong to a similar “kernel” that requires unique knowledge for identification. Thus, akin to the pedagogical principles in human education, the optimal approach for expanding knowledge in the SEA-net-based system involves prioritizing the acquisition of a primary set of symbols.

\textbf{Knowledge transfer between separate systems through symbolic communication}

Language is one of the most efficient mechanisms for transferring knowledge among humans (1, 24) and can reduce or even eliminate costly trial-and-error learning. Thus, we next examined whether the symbols that emerged in SEA-net can support knowledge transfer based on communication between separate systems. To this end, we conducted a “learning-by-communication” game, in which one agent trained to recognize a specific category can “tell” another agent to perform the same task through communication (Fig. 3a). Here, two agents were trained (as illustrated in Fig. 1) independently and derived their own symbol sets separately. For communication analysis, the two symbol systems were first aligned with the introduction of two additional modules, i.e., translating-in (TI) and translating-out (TO) modules (Fig. 3b, upper panel), with the symbol set that emerged in one agent mapped to that of the other. Under the two-agent scenario, we assigned “speaker” and “listener” roles. The TO module of the speaker agent was omitted for simplicity (Fig. 3b, lower panel). In the message alignment stage, the TI module was used to map the symbols generated by the speaker agent to those generated by the listener agent. Similar to the experiments described in Fig. 3, the speaker agent was trained using all 100 classes, while the listener agent was trained using 99 classes. After completing TI module training on the symbols of the overlapping 99 classes, the listener agent was challenged to recognize an untrained category based on instructions given by the (translated) symbols from the speaker agent.

The task was performed for 100 rounds. In each round, a different category was designated as the untrained testing class for the listener agent. As shown in Fig. 3c, the listener agent performed quite well in most testing classes, validating the effectiveness of the emerged symbols in knowledge transfer. Interestingly, the performance of this “learning-by-communication” game across categories was significantly correlated with that of the symbolic inferring experiment (Fig. 3d). These results imply that when two symbol sets emerge in distinct agents, they exhibit a shared structure of compositional semantics, enabling the SEA-net-based system to gain new knowledge without the need for explicit learning.

\textbf{Emerged symbols in SEA-net resemble the semantic structure of natural language}

Comparing the self-organized SEA-net symbol system with natural human language may help identify common mechanisms underlying the emergence of symbols. To this end, we first visualized the similarity matrix among the word vectors of category names (e.g., “boy”, “bee”) according to the order of categories obtained by clustering analyses of SEA-net-generated symbols (Fig. 2a). If clustering in this similarity matrix similar to Fig. 2a, it would imply shared semantic information contained in the symbols generated by SEA-net and natural language. Fig. 4a shows the result for the word vectors. We can see several clusters similar to Fig. 2a. Notably, overlapping clusters between the SEA-net-generated symbols and natural language category names included people (e.g., woman, man, girl, boy, baby), animals (e.g., bear, elephant, and cattle), transportation (e.g., bicycle, streetcar, train, bus, and motorcycle), and landscapes (e.g., mountain and sea) (red squares in Fig. 4a). To quantify the similarity between the semantic structures of the two symbol systems, we calculated the correlation coefficients between the dendrogram of the SEA-net-generated symbols and the cosine distance matrix of natural language category-name word vectors (43), yielding a significantly greater than chance result (0.341) (Fig. 4c). Overall, these findings indicated a similar semantic structure obtained by SEA-net and natural language, suggesting a possible general mechanism through which meanings can be distilled into symbols.

Finally, to determine whether the SEA-net network architecture can exploit the semantics embedded in natural language, we assessed whether natural language can be directly used for knowledge transfer in the “learning-by-communication” game. Specifically, individual class names, represented by corresponding word vectors, were reduced to the same size as the SEA-net-generated symbols and then directly fed into the CDP module. Only the neural network parameters were updated during the training, while the name-based symbolic inputs were kept fixed. The agent was trained using 99 classes with corresponding names, then challenged to recognize the untrained class with its name as the symbolic input. As shown in Fig. 4b, the system well recognized the images for most classes, even though the system never encountered the images or names before the test. These results suggest that the framework proposed here, similar to previous networks recommended for zero-shot learning (25), can effectively exploit the compositional semantics embedded in natural language to solve new problems. In addition, the accuracies across categories were highly correlated in the two “learning-by-communication” experiments (comparing Fig. 3c and Fig. 4c, also see Fig. 4D), confirming the existence of a common semantic structure captured by symbols generated by SEA-net and natural language.

\textbf{Discussion}

The ability to understand semantics is considered important for achieving human-level intelligence (8–11). We propose a solution that allows the emergence of symbols and grounds their meaning in artificial neural networks. We propose that the meanings should be grounded on the basis of a connectionist system, i.e., the network structure itself. While the grounding process was demonstrated here using a visual identification task, the framework could be readily extended to other input modalities and tasks. Indeed, any function that a network can perform, such as identification of an object, execution of a motor command, or comprehension of an abstract concept, could serve as the ground upon which a symbol obtains its meaning. Our proposal posits that the internal structure within the human brain may potentially serve as the equivalent of Chomsky's "deep structure" of language (26, 27).

Although the described process of symbol emergence is guided by supervised learning with categories defined by humans, it is readily applicable in reinforcement learning settings, enabling an agent to autonomously formulate concepts and generate corresponding symbols without external guidance. Thus, the meaning of the symbols will not be constrained by natural human language semantics but will instead reflect both the structural properties of the network itself and the nature of its interaction with the environment. Humans are highly efficient in forming new symbols, either to represent new concepts or to communicate among individuals without a common language (28). The ability to create meaningful symbols, rather than relying on existing ones, is at the core of human language faculty and is the very reason why language can continuously evolve to fit human conditions. In this regard, SEA-net suggests a framework capable of open-ended symbol generation and understanding that can better support dynamically changing AI conditions.

The SEA-net framework also answers the profound question raised in Chinese Room argument (CRA): i.e., what does semantic understanding mean for an AI system and even for humans (29)? Given the results presented here, we suggest that semantic understanding is equivalent to an internalized operation. Thus, for a neural network, understanding a symbol means being able to configure or choose a specific structure to perform certain functions referred to by this symbol. For example, within the visual identification domain, an agent can be regarded as understanding the meaning of a symbol representing an apple if the symbol can configure a specific structure to recognize the apple from alternative categories. This view of understanding is consistent with the EC (11, 30) and GC concepts (6, 12). Importantly, SEA-net provides a practical framework to implement EC in artificial neural networks. In accordance with the EC perspective, the symbols generated by SEA-net exhibit a close relationship with the structure of each individual network, thereby potentially exhibiting variations across different networks. Nevertheless, despite the individualized nature of the symbols that refer to the same referent, they can be mapped onto an arbitrarily defined common symbol to facilitate communication, mirroring the process by which humans acquire the ability to speak a shared language. 

The SEA-net architecture is a promising candidate for future models, combining symbolic representation and manipulation with a connectionist structure (8, 31–33). Recent advances in the field of natural language processing demonstrate the impressive ability of deep networks to process syntactic information, which deals with complicated sequences of words. In the future, adding a syntax processing (SP) module to the CDP module would be important, enabling SEA-net to handle sophisticated symbol sequences following specific syntactic rules. If this can be achieved, the enhanced CDP module may dynamically configure the TS module to execute a series of operations that simulate specific scenarios without them occurring in the real world, similar to the concept of “embodied simulation” (12). This would endow the network with functions such as planning and reasoning, which are currently challenging for AI.

Current deep neural networks rely on structural changes to acquire new functions. However, as suggested by the SEA-net architecture, it is possible to achieve the same purpose by dynamically choosing a proper configuration without structural changes. This framework can capitalize on the power of composability by achieving almost infinite functions with diverse but limited structural motifs. Furthermore, the SEA-net architecture exhibits the potential for recursive application. At the lower level, a CDP or SP module possesses its own network structure, which can be considered as a TS module when employed by another CDP or SP module at a higher level.. However, further research is needed to investigate the scalability of this recursive structured system and to ascertain whether it has the potential to eventually mimic one of the most remarkable characteristics of human language: the ability to compose and comprehend boundless internal expressions despite limited resources (14).

The SEA-net architecture proposed here is inspired by the semantic network of the human brain. Similar to the separation of the CDP and TS modules in our proposal, distinct sub-networks in the brain are responsible for conceptual and sensory-motor processing (34, 35), both of which are involved in semantic processing (34, 36–38). Specifically, the PFC, which inspired the CDP module, is essential for forming concepts (39). In addition, similar to the TS module, language comprehension depends on a widespread sensory-motor network (38), e.g., the involvement of the primary motor and premotor cortices in processing action verbs and higher-level visual cortices in processing object nouns (34). Moreover, the gating operation that links information processed by the CDP and TS modules is biologically plausible (21). Notably, several synaptic mechanisms have been proposed as efficient means for neuronal gain control (40, 41), resembling the gating operation in SEA-net. It would be informative to investigate whether the interaction between the PFC and sensory-motor areas in the brain is indeed mediated through a mechanism like gain control. These findings may shed new light on our understanding of the relationship between language and the brain. Based on the current results, the composability of human language may be deeply rooted in the composability of the myriad of functional neural circuits in the brain. Thus, language can be considered as an abstract representation of the dynamics in the brain and studying the hierarchical structure of language may reveal deep mysteries about how our brain is functionally organized at various scales.

\textbf{Methods}

\textbf{The structure of networks used}

\emph{\textbf{SEA-net}.} As illustrated in Fig. 1A, SEA-net consists of two parts: the CDP module and the TS module. The TS module can exhibit various network structures depending on the task. The CDP module, inspired by the structure of the prefrontal cortex (PFC), transforms symbolic input into proper controlling signals for manipulating the processing of the TS module (\emph{19}). As shown in Fig. 1A, the $k^{\text{th }}$ neuron in the $l^{\text{th. }}$ layer in the CDP module generates a controlling signal $g_k^l$, and sends it to the $k^{\text{th }}$. neuron in the corresponding layer $m$ in the TS module, which modulates its output $y_k^m$ to $z_{k}^{m}=g_{k}^{l}y_{k}^{m}$.In this framework, the controlling neuron in the CDP module is one-to-one matched to the controlled neuron in the TS module. Thus, the $l^{\text{th }}$layer of CDP module possesses the same number of neurons as the ${{m}^{th}}$ layer of the TS module.

In the present work, the TS module includes two parts: a feature extractor and a classifier. We employed a pretrained convolutional neural network (CNN), i.e., the ResNet18 module from Torchvision library (\emph{42}), as the feature extractor. The CDP module did not modulate the feature extractor in the experiments mentioned in the main text. However, it is possible to utilize CDP signals in the same way, to modulate the processing of the feature extractor. The feature extractor outputted the features to the classifier through an identity layer. The classifier consisted of a three-layer perceptron with {[}512-100-10{]} neurons using the ReLU activation function. The output layer of the classifier consisted of two units assigned with a cross-entropy loss and representing ``Yes'' {[}0 1{]} and ``No'' {[}1 0{]}, respectively. As a match with the structure of the classifier in the TS module, the CDP module also consisted of three fully connected layers of {[}512-100-10{]} units. The activation function for neurons in the CDP module is sigmoid, which generates the output controlling signals ranging from 0 to 1. We also tested the efficacy of the SEA-net framework with other network specifications. For example, experiments similar to that shown in Fig. 1D were conducted with a TS module consisting of a feature extractor of ResNet34 followed by a classifier of {[}512-100-100-100{]} fully-connected layers, and the CDP module consisting of six fully-connected layers of {[}200-200-512-100-100-100{]} neurons, with the last four layers outputting controlling signals to the classifier in the TS module. In this experiment, the feature extractor was trained together with the rest of the network during the learning process. The results are shown in Fig. S3, indicating that the SEA-net framework can be effectively applied to various network structures.

\textbf{\emph{TI module.}} The TI module (Fig. 3B) translates the symbols generated by one agent to that of another agent. In all experiments, the TI module used was a multiple-layer perceptron, with ten hidden layers containing 500 neurons each. The ReLU activation function and the mean squared error (MSE) loss function were used. During the training, the dropout probability of all hidden layers was set to 0.3.

\textbf{Training of SEA-net}

At the beginning of the training, a real vector randomly chosen for each image category was fed to the CDP module as the initial for the self-generated symbol. In the training process, false symbols, i.e., an image from class A assigned with the symbol representing class B, were given with a possibility \(p\) to construct negative samples. \(p\) randomly varied between 0 and 1 from batch to batch to facilitate learning. Depending on whether an image fed into the TS module was matched to its corresponding symbol, the supervising signal was chosen to be ``yes'' or ``no''. The training process was divided into two phases, as illustrated in Fig. 1C. The symbols and network parameters were modified in the symbol training phase and network training phase, respectively. The feedforward propagation of information in SEA-net was the same in the two training phases. In the network training phase, only parameters of the network, including both the CDP and trainable part of the TS modules, were updated according to the backpropagated gradients \(\Delta W\). In the symbol training phase, the gradients \(\Delta C\) backpropagated through the classifier of the TS module and the CDP module modified the input symbols, while all network structure parameters were kept fixed. The two phases in the training process were carried out alternatively in an epoch-by-epoch manner. The training was terminated after 2000 epochs to ensure accuracy reaching the plateau.

In the experiments with predefined symbols, e.g., when the word vector of category name was used, only the network training phase was involved. In both conditions of self-generated and predefined symbols, uniformly distributed noise ranging from -0.1 to 0.1 was injected into each element of the symbols independently in both the network training phase and symbol training phase (if any), except for the symbolic manipulation experiments. We found that it effectively increased the system's robustness for distinguishing various categories. No noise was added to the symbols in the testing. In all experiments, the length of symbols was set to 20, i.e., they contained 20 real elements. The SEA-net, including the symbols and the network parameters, was trained with a learning rate of 0.0001. In the experiment shown in Fig. S3, the learning rate of ResNet34 was set to 0.000001, while the rest hyper-parameters remained the same.

\textbf{Learning by symbolic manipulation}

In these experiments, firstly, the network was trained, as described above, by dataset

$D_{99}$ containing 99 classes. Then the task of identifying the remaining class $D_1$ was trained by symbolic manipulation only, through the symbol training phase, as illustrated in Fig. 1C. To utilize the knowledge obtained by-learning to identify. $D_w$ as much as possible, we introduced a repelling loss $L_{rep}$ for learning symbolic manipulation, which was defined as

\[
L_{rep}=\sum_{i \in I_t} \exp \left(-\textbar S_i-S \textbar^{2} / \tau\right) \tag{1}
\]

where $S_i, i \in I_t$ were the symbols of classes belonging to $D_{99}$ and $S$ the symbols of the remaining class in $D_1$. To test the system's capability of small sample learning, only two images belonging to $D_1$ and one image from each of the 99 learned classes belonging to $D_{99}$ were utilized in training. The symbols assigned to the class in $\mathrm{D}_1$ were randomly initialized and trained to minimalize the following loss function:

\[
L={{L}_{CE}}\left( {{x}_{\text{new }}},y\mid S \right)+\alpha {{L}_{CE}}\left( {{x}_{\text{old }}},\bar{y}\mid S \right)+\beta {{L}_{\text{rep }}}\left( {{S}_{\text{new }}},{{S}_{\text{old }}},\tau \right) \tag{2}
\]

where $L_{CE}$ denotes the cross-entropy loss, $L_{rep}$ a the repelling loss defined in equation (1), $x_{\text{new }}$ the image sample from the new class in ${{D}_{1}}$, ${{x}_{\text{old }}}$ the image sample from the learned classes in $D_{99},$ $y$ the label "Yes", $\bar{y}$ the label "No" and $\alpha, \beta$ are parameters used to balance the different contributions of the losses.. Hyper-parameters in these experiments were set to $\alpha=0.5, \beta=0.001, \tau=0.01$, and the learning rate to 0.01. The accuracy was tested on the CIFAR 100 testing set, with 100 images of the new class mixed with 2000 images of the learned class. The experiment was iterated on each category in the CIFAR 100 as the new class learned by symbolic manipulation. In each experiment, ten realizations were conducted with randomly selected $\{x_{old}, x_{new}\}$ image combinations.

\textbf{Training the TI module}

The TI module was trained to map the symbols generated by the speaker agent to that generated by the listener agent. Firstly, according to the procedure for training SEA-net, the speaker agent generated one symbol for each category in $D$ $(D={{D}_{99}}\cup {{D}_{1}})$, and the listener agent generated one symbol for each category in $D_{99}$\textsubscript{.} To generate enough samples for training, the speaker symbol dataset was extended to 97 symbols for each category by symbolic manipulation. Specifically, after the initial training of the speaker agent as described in the section ``training of SEA-net'', the network parameters were fixed. Then 96 additional symbols for each category were obtained through the training procedure described in the ``learning by symbolic manipulation'' section. Each symbol was obtained by training 1000 epochs starting from different random initial symbols. TI was trained, for $D_{99}$, to map the 97 symbol sets for the speaker agent to the corresponding 1 set of symbols for the listener agent. The learning rate was set to 0.0001 and decayed by the factor of 0.5 for every ten epochs.~The Adam algorithm (\emph{43}) was used. The training lasted for 200 epochs to ensure convergence.

\textbf{The procedure of the ``learning-by-communication'' game}

The communication experiment included two stages: 1) message alignment and 2) identification. In the stage of message alignment (cf.~the upper panel in Fig. 3A), two SEA-net agents were trained separately and independently generated their own symbol set. The speaker agent was trained with dataset $D$ of all 100 classes, while the listener agent was trained with $D_{99}$ containing 99 classes. The TI module of the listener agent was then trained to map the two symbol sets for $D_{99}$. After successfully training the TI module, we moved to the identification (cf.~the lower panel in Fig. 3A).

In this stage, the symbol generated by the speaker agent corresponding to the remaining untrained class for the listener agent, $D_1$, was translated by the TI module and then fed into the CDP module of the listener agent, which instructed the listener agent to identify the correct images. The experiment was repeated 100 rounds, with a different class chosen as $D_1$ in each round. As a control experiment, we also tested the accuracy of the learning-by-communication game with random symbols used for the listener agent, which were uniformly sampled ranging from the lower bound to the upper bound of self-generated symbols of the listener agent.

\textbf{Training SEA-net with natural language}

In these experiments, SEA-net was trained using the category-name word vectors (43) as the predefined symbols. The dataset was divided into two parts, $D_{99}$ and $D_1$, in the same way as in the ``learning-by-communication'' game. SEA-net was directly trained by class names, represented by their word vectors, with images belonging to $D_{99}$. Then it was tested with the untrained class name corresponding to $D_1$ to identify the correct images. Similar to the ``learning-by-communication'' game, experiments were repeated 100 rounds with each class chosen as $D_1$.

\textbf{Clustering analysis of generated symbols}

We used agglomerative clustering (Matlab function \emph{dendrogram}) to group symbols generated by SEA-net, based on cosine distance between symbols and unweighted average linkage between clusters. The dendrogram, a tree-based representation, demonstrated the clustering-analysis results reported in Fig. 2A. The semantic network of symbols in Fig. 2B was calculated based on the clustering results. Specifically, two symbols, each from one of two distinct but connected branches or leaves in the dendrogram with the closest distance, were connected by one edge. We traversed all pairs of connected branches and leaves, linking all pairs of symbol nodes to meet the requirement of the closest distance. The visualization of the semantic network was generated by Gephi (\emph{44}).

\textbf{Cophenetic correlation coefficient calculation}

An index derived from the cophenetic correlation coefficient was used to measure the similarity between the internal structure of symbols generated by SEA-net and that of the word vectors of class names. Specifically, the internal structure of symbols generated by SEA-net was represented by the dendrogrammatic distance $t_{i j}$ between the leaf categories $i$ and $j$, defined as the node height at which the two leaves were first joined together. The cosine distance $d_{i j}$ represented the internal structure of the word vectors of class names between the categories $i$ and $j$. Then, their cophenetic correlation coefficient (\emph{45}) $c(t, d)$ was defined as

\[
\quad c(t, d)=\frac{\sum_{i\textless j}\left(d^X(i, j)-\bar{d}^X\right)\left(t^Y(i, j)-\bar{t}^Y\right)}{\sqrt{\left[\sum_{i\textless j}\left(d^X(i, j)-\bar{d}^X\right)^2\right]\left[\sum_{i\textless j}\left(t^Y(i, j)-\bar{t}^{\mathrm{Y}}\right)^2\right]}} \quad \tag{3}
\]

where $\bar{d}^X$. is the mean of the $d_{i, j}^X$, and $\bar{t}^Y$ the mean of the $t_{i, j}^Y$ To examine the statistical significance of the obtained cophenetic correlation coefficient, we shuffled the elements of symbols and recalculated the coefficient 1000 times.

\textbf{Datasets}

We used CIFAR 100 (\emph{46}) images as the sensory inputs of SEA-net in all experiments. CIFAR 100 dataset contains 50,000 training images and 10,000 testing images (32×32 pixels with color), containing 100 fine-grained classes.

The predefined word vectors of class names were provided by the fastText library(\emph{47}). Specifically, the dataset contains 2 million word vectors with a length of 300. The word vectors of fine-grained class names in CIFAR 100 extracted from the dataset were reduced to the length of 20 by the tools provided by the library. The magnitude of each vector was amplified ten times before being fed into the CDP module.

\textbf{References}

1. T. W. Deacon, \emph{The Symbolic Species: The Co-evolution of Language and the Brain} (W. W. Norton \& Company, 1998).

2. A. Newell, \emph{Unified Theories of Cognition} (Harvard University Press, 1994).

3. K. Laland, A. Seed, Understanding Human Cognitive Uniqueness. \emph{Annu. Rev.~Psychol.} \textbf{72}, 689--716 (2021).

4. S. Bubeck, V. Chandrasekaran, R. Eldan, J. Gehrke, E. Horvitz, E. Kamar, P. Lee, Y. T. Lee, Y. Li, S. Lundberg, H. Nori, H. Palangi, M. T. Ribeiro, Y. Zhang, Sparks of Artificial General Intelligence: Early experiments with GPT-4 (2023), , doi:10.48550/arXiv.2303.12712.

5. R. Bommasani, D. A. Hudson, E. Adeli, R. Altman, S. Arora, S. v       on Arx, M. S. Bernstein, J. Bohg, A. Bosselut, E. Brunskill, E. Brynjolfsson
% ...
On the Opportunities and Risks of Foundation Models. \emph{arXiv:2108.07258 {[}cs{]}} (2021) (available at http://arxiv.org/abs/2108.07258).

6. B. K. Bergen, \emph{Louder Than Words: The New Science of How the Mind Makes Meaning} (Basic Books, 2012).

7. T. Taniguchi, E. Ugur, M. Hoffmann, L. Jamone, T. Nagai, B. Rosman, T. Matsuka, N. Iwahashi, E. Oztop, J. Piater, F. Wörgötter, Symbol Emergence in Cognitive Developmental Systems: A Survey. \emph{IEEE Transactions on Cognitive and Developmental Systems}. \textbf{11}, 494--516 (2019).

8. P. Hitzler, A. Eberhart, M. Ebrahimi, M. K. Sarker, L. Zhou, Neuro-Symbolic Approaches in Artificial Intelligence. \emph{National Science Review}, nwac035 (2022).

9. R. Pfeifer, M. Lungarella, F. Iida, Self-Organization, Embodiment, and Biologically Inspired Robotics. \emph{Science}. \textbf{318}, 1088--1093 (2007).

10. M. L. Anderson, Embodied Cognition: A field guide. \emph{Artificial Intelligence}. \textbf{149}, 91--130 (2003).

11. A. Cangelosi, J. Bongard, M. H. Fischer, S. Nolfi, "Embodied Intelligence" in \emph{Springer Handbook of Computational Intelligence}, J. Kacprzyk, W. Pedrycz, Eds. (Springer, Berlin, Heidelberg, 2015; https://doi.org/10.1007/978-3-662-43505-2\_37), pp.~697--714.

12. L. W. Barsalou, Grounded Cognition. \emph{Annu. Rev.~Psychol.} \textbf{59}, 617--645 (2008).

13. S. Harnad, The symbol grounding problem. \emph{Physica D: Nonlinear Phenomena}. \textbf{42}, 335--346 (1990).

14. M. D. Hauser, N. Chomsky, W. T. Fitch, The Faculty of Language: What Is It, Who Has It, and How Did It Evolve? \emph{Science}. \textbf{298}, 1569--1579 (2002).

15. P. Liang, C. Potts, Bringing Machine Learning and Compositional Semantics Together. \emph{Annu. Rev.~Linguist.} \textbf{1}, 355--376 (2015).

16. M. Taddeo, L. Floridi, Solving the symbol grounding problem: a critical review of fifteen years of research. \emph{Journal of Experimental \& Theoretical Artificial Intelligence}. \textbf{17}, 419--445 (2005).

17. P. Vogt, The physical symbol grounding problem. \emph{Cognitive Systems Research}. \textbf{3}, 429--457 (2002).

18. L. Steels, M. Hild, \emph{Language Grounding in Robots} (Springer Science \& Business Media, 2012).

19. G. Zeng, Y. Chen, B. Cui, S. Yu, Continual learning of context-dependent processing in neural networks. \emph{Nature Machine Intelligence}. \textbf{1}, 364--372 (2019).

20. J. Russin, R. C. O'Reilly, Y. Bengio, Deep learning needs a prefrontal cortex. \emph{Work Bridging AI Cogn Sci}. \textbf{107}, 603--616 (2020).

21. B. Tsuda, K. M. Tye, H. T. Siegelmann, T. J. Sejnowski, A modeling framework for adaptive lifelong learning with transfer and savings through gating in the prefrontal cortex. \emph{Proceedings of the National Academy of Sciences}. \textbf{117}, 29872--29882 (2020).

22. F. Lehmann, Semantic networks. \emph{Computers \& Mathematics with Applications}. \textbf{23}, 1--50 (2002).

23. P. Vincent-Lamarre, A. B. Massé, M. Lopes, M. Lord, O. Marcotte, S. Harnad, The Latent Structure of Dictionaries. \emph{Top Cogn Sci}. \textbf{8}, 625--659 (2016).

24. A. Blondin-Massé, S. Harnad, B. St-Louis, "Symbol Grounding and the Origin of Language: From Show to Tell" in, C. Lefebvre, H. Cohen, B. Comrie, Eds. (Benjamin, 2013; https://eprints.soton.ac.uk/271438/).

25. A. Radford, J. W. Kim, C. Hallacy, A. Ramesh, G. Goh, S. Agarwal, G. Sastry, A. Askell, P. Mishkin, J. Clark, G. Krueger, I. Sutskever, "Learning Transferable Visual Models From Natural Language Supervision" in \emph{Proceedings of the 38th International Conference on Machine Learning} (PMLR, 2021; https://proceedings.mlr.press/v139/radford21a.html), pp.~8748--8763.

26. N. Chomsky, \emph{Rules and Representations} (Columbia University Press, 2005).

27. A. Djeribiai, Chomsky's Generative Transformational Grammar and its Implications on Language Teaching (2016) (available at http://dspace.univ-eloued.dz:80/xmlui/handle/123456789/2771).

28. M. H. Christiansen, N. Chater, \emph{The Language Game: How Improvisation Created Language and Changed the World} (Hachette UK, 2022).

29. J. R. Searle, Minds, brains, and programs. \emph{Behavioral and Brain Sciences}. \textbf{3}, 417--424 (1980).

30. J. Feldman, S. Narayanan, Embodied meaning in a neural theory of language. \emph{Brain and Language}. \textbf{89}, 385--392 (2004).

31. G. Marcus, The Next Decade in AI: Four Steps Towards Robust Artificial Intelligence. \emph{arXiv:2002.06177 {[}cs{]}} (2020) (available at http://arxiv.org/abs/2002.06177).

32. A. d'Avila Garcez, L. C. Lamb, Neurosymbolic AI: The 3rd Wave. \emph{arXiv:2012.05876 {[}cs{]}} (2020) (available at http://arxiv.org/abs/2012.05876).

33. R. Sun, F. Alexandre, Eds., \emph{Connectionist-Symbolic Integration: From Unified to Hybrid Approaches} (Psychology Press, New York, 1997).

34. J. R. Binder, R. H. Desai, W. W. Graves, L. L. Conant, Where Is the Semantic System? A Critical Review and Meta-Analysis of 120 Functional Neuroimaging Studies. \emph{Cerebral Cortex}. \textbf{19}, 2767--2796 (2009).

35. J. A. Fodor, \emph{The Modularity of Mind} (MIT Press, 1983).

36. P. T. Schoenemann, "Chapter 22 - Evolution of brain and language" in \emph{Progress in Brain Research}, M. A. Hofman, D. Falk, Eds. (Elsevier, 2012; https://www.sciencedirect.com/science/article/pii/B9780444538604000222), vol.~195 of \emph{Evolution of the Primate Brain}, pp.~443--459.

37. J. D. E. Gabrieli, R. A. Poldrack, J. E. Desmond, The role of left prefrontal cortex in language and memory.\emph{Proceedings of the National Academy of Sciences}. \textbf{95}, 906--913 (1998).

38. A. G. Huth, W. A. de Heer, T. L. Griffiths, F. E. Theunissen, J. L. Gallant, Natural speech reveals the semantic maps that tile human cerebral cortex. \emph{Nature}. \textbf{532}, 453--458 (2016).

39. E. K. Miller, J. D. Cohen, An Integrative Theory of Prefrontal Cortex Function. \emph{Annual Review of Neuroscience}. \textbf{24}, 167--202 (2001).

40. L. F. Abbott, F. S. Chance, "Drivers and modulators from push-pull and balanced synaptic input" in \emph{Progress in Brain Research} (Elsevier, 2005; https://www.sciencedirect.com/science/article/pii/S0079612305490111), vol.~149 of \emph{Cortical Function: a View from the Thalamus}, pp.~147--155.

41. L. F. Abbott, J. A. Varela, K. Sen, S. B. Nelson, Synaptic Depression and Cortical Gain Control. \emph{Science}. \textbf{275}, 221--224 (1997).

42. K. He, X. Zhang, S. Ren, J. Sun, "Deep Residual Learning for Image Recognition" in (2016; CVPR), pp.~770--778.

43. Kingma, D.P., Ba, L.J., Amsterdam Machine Learning lab (IVI, FNWI), "Adam: A Method for Stochastic Optimization" in \emph{International Conference on Learning Representations (ICLR)} (arXiv.org, 2015; https://dare.uva.nl/personal/pure/en/publications/adam-a-method-for-stochastic-optimization(a20791d3-1aff-464a-8544-268383c33a75).html).

44. M. Bastian, S. Heymann, M. Jacomy, Gephi: An Open Source Software for Exploring and Manipulating Networks. \emph{Proceedings of the International AAAI Conference on Web and Social Media}. \textbf{3}, 361--362 (2009).

45. S. Saraçli, N. Doğan, İ. Doğan, Comparison of hierarchical cluster analysis methods by cophenetic correlation. \emph{Journal of Inequalities and Applications}. \textbf{2013}, 203 (2013).

46. A. Krizhevsky, G. Hinton, Learning multiple layers of features from tiny images (2009).

47. A. Joulin, E. Grave, P. Bojanowski, M. Douze, H. Jégou, T. Mikolov, ``FastText.zip: Compressing text classification models'' (arXiv:1612.03651, arXiv, 2016), , doi:10.48550/arXiv.1612.03651.

48. P.-S. Huang, X. He, J. Gao, L. Deng, A. Acero, L. Heck, "Learning deep structured semantic models for web search using clickthrough data" in \emph{Proceedings of the 22nd ACM international conference on Conference on information \& knowledge management - CIKM '13} (Association for Computing Machinery, New York, NY, USA, 2013; http://dl.acm.org/citation.cfm?doid=2505515.2505665), \emph{CIKM '13}, pp.~2333--2338.

\textbf{Acknowledgments:}

The authors thank Danko Nikolić, Frederic Alexandre and Jinpeng Zhang for helpful comments and discussions.

\textbf{Funding:}

This work was supported by grants from:

International Partnership Program of Chinese Academy of Sciences 173211KYSB2020002 (S.Y)

CAS Project for Young Scientists in Basic Research YSBR-041 (Y. C)

The Strategic Priority Research Program B of the Chinese Academy of Sciences XDB32040201 (S.Y.)

National Natural Science Foundation of China 11905291 (Y.C)

\textbf{\hfill\break
}

\textbf{Figures}

\begin{figure}
\centering
\includegraphics[width=4.9in,height=4.06857in]{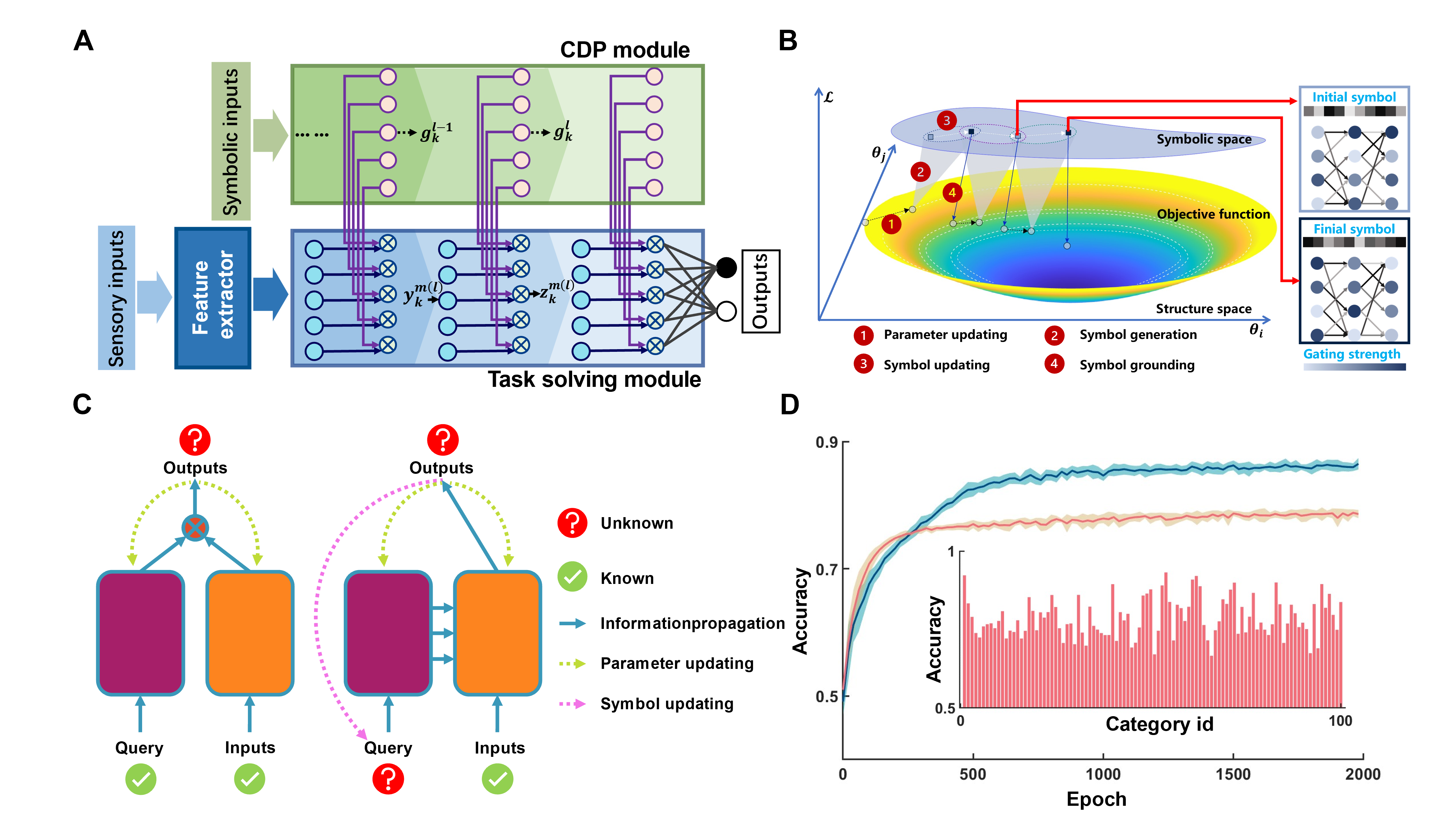} 
\caption{\textbf{Network structure and training process of SEA-net.}
\textbf{(A)} SEA-net structure. System backbone contains CDP (green) and TS modules (blue). TS module incepts sensory inputs and outputs proper classification results. CDP module uses symbols to modulate TS module processing.
\textbf{(B)} Diagram of the two-phase training process. In the network training phase (with steps \ding{182}\ding{185}), all network parameters are updated according to the output layer's backpropagated gradients with fixed symbols. In the symbol training phase (with steps \ding{183}\ding{184}), only the symbols are modified while all network parameters are kept fixed. Modified symbols will assign new configurations to the TS module.
\textbf{(C)} Comparison of SEA-net with two-tower models (25, 48). Within the two-tower model framework, a predefined set of queries and an input stream are utilized, and both modules aim to transform them into proper embedding by maximizing their inner product. In SEA-net, only inputs are predefined, and queries need to be determined to select substructures in the TS module to generate proper responses during learning. SEA-net is trained on the parameter updating phase during an epoch and symbol updating phase during the next epoch.  \textbf{(D)} SEA-net performance during the training process. Green line represents accuracy during training, while orange line represents accuracy during testing. Inset: testing accuracy of each class of CIFAR 100. See Table S1 for category names.}
\end{figure}

\begin{figure}
\centering
\includegraphics[width=4.9in,height=7.8in]{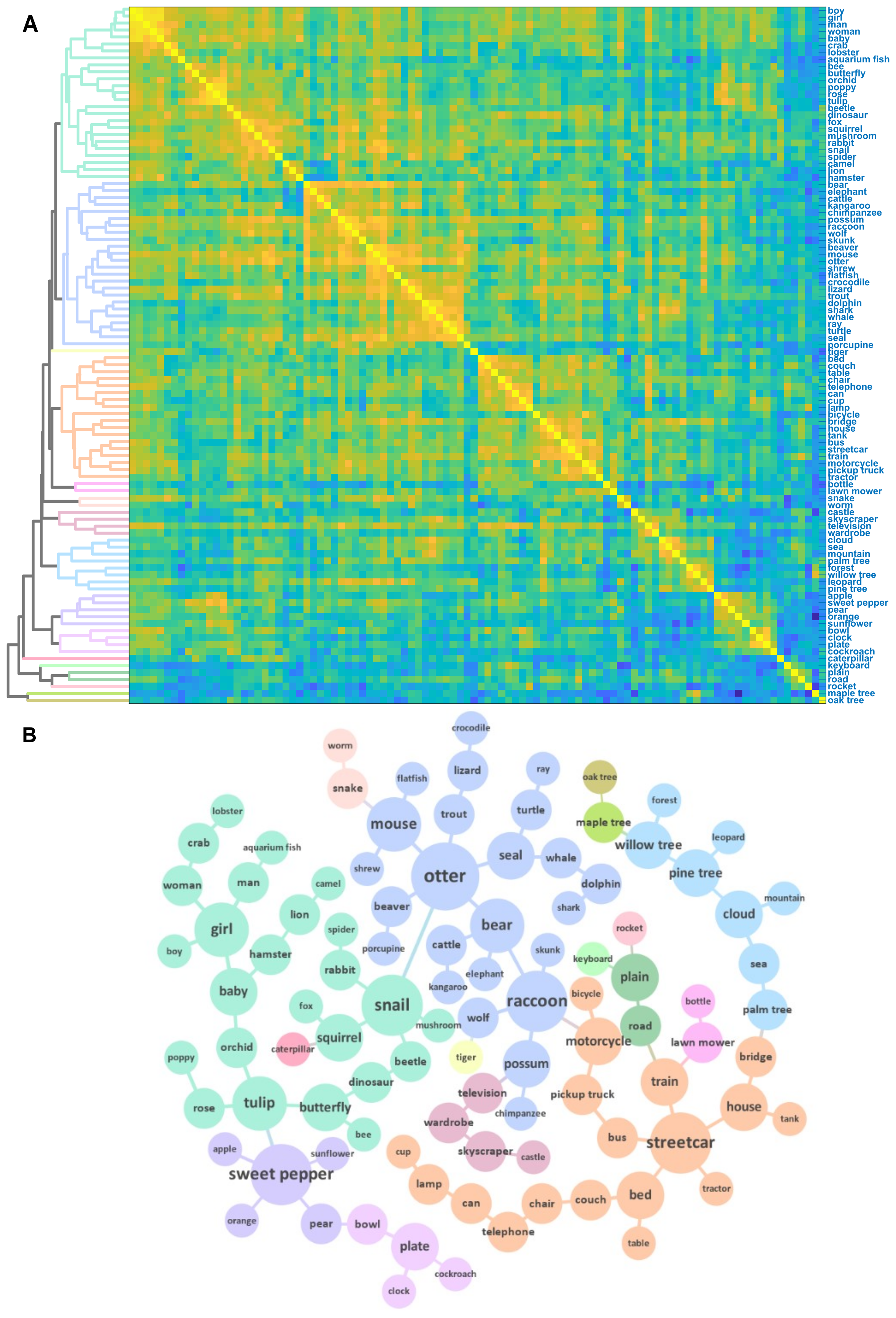}
\captionsetup{labelformat=empty}
\caption{}
\end{figure}
\begin{figure}
\ContinuedFloat
\caption{ \textbf{ Clustering analysis and semantic network of generated symbols.} 
\textbf{(A)} Cosine similarity matrix of generated symbols, with index order rearranged according to hierarchical clustering results. Corresponding dendrogram is shown on the left of the matrix. Labels of each class are shown on the right of the matrix.  
\textbf{(B)} Semantic network of generated symbols. Nodes of symbols belonging to the same cluster are represented by the same color and correspond to that in the dendrogram (A). Size of node represents degree, i.e., number of connections to a node, in the network.}
\end{figure}

\begin{figure}
\centering
\includegraphics[width=4.9in,height=4.31875in]{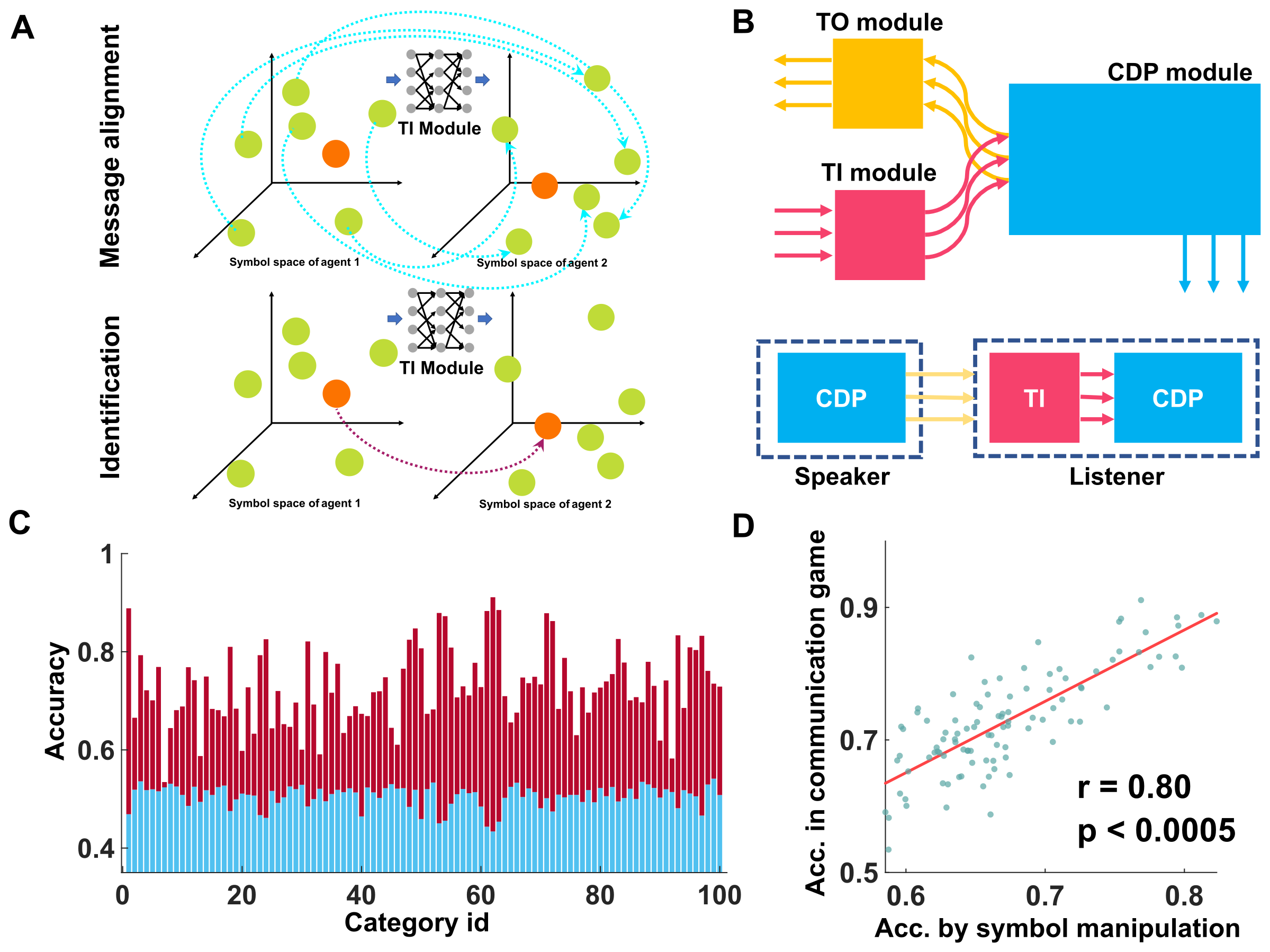}
\caption{ \textbf{Communications between two SEA-net-based agents.} 
\textbf{(A)} Communication game diagram. The game consists of two stages: message alignment and identification. In message alignment, two independently trained agents learn to match their acquired symbols (green dots in symbol space) for the same set of classes. In the identification stage, the speaker agent outputs a symbol (orange dots in symbol space) corresponding to one class, and the listener agent needs to translate the symbol into its own representation and judge whether an image belongs to the class based on it. 
\textbf{(B)} Auxiliary modules of SEA-net for communication. TI module translates public symbols to the agent’s own symbol system, and the TO module translates the agent’s own symbols to public symbols (upper diagram). As we assigned the symbols generated by the speaker agent as public symbols, only the TI module was used for communication (lower diagram). 
\textbf{(C)} Identification accuracy (dark red bar) of the listener agent on new classes, instructed by the (translated) symbols provided by the speaker agent. Light blue bar represents accuracy achieved with random symbols. 
\textbf{(D)} Scatter plot of accuracy for individual classes in the communication game shown in (C) vs. that achieved by symbolic manipulation shown in Fig. S1. Red line: linear regression.}
\end{figure}

\begin{figure}
	\centering
	\includegraphics[width=4.9in,height=7.8in]{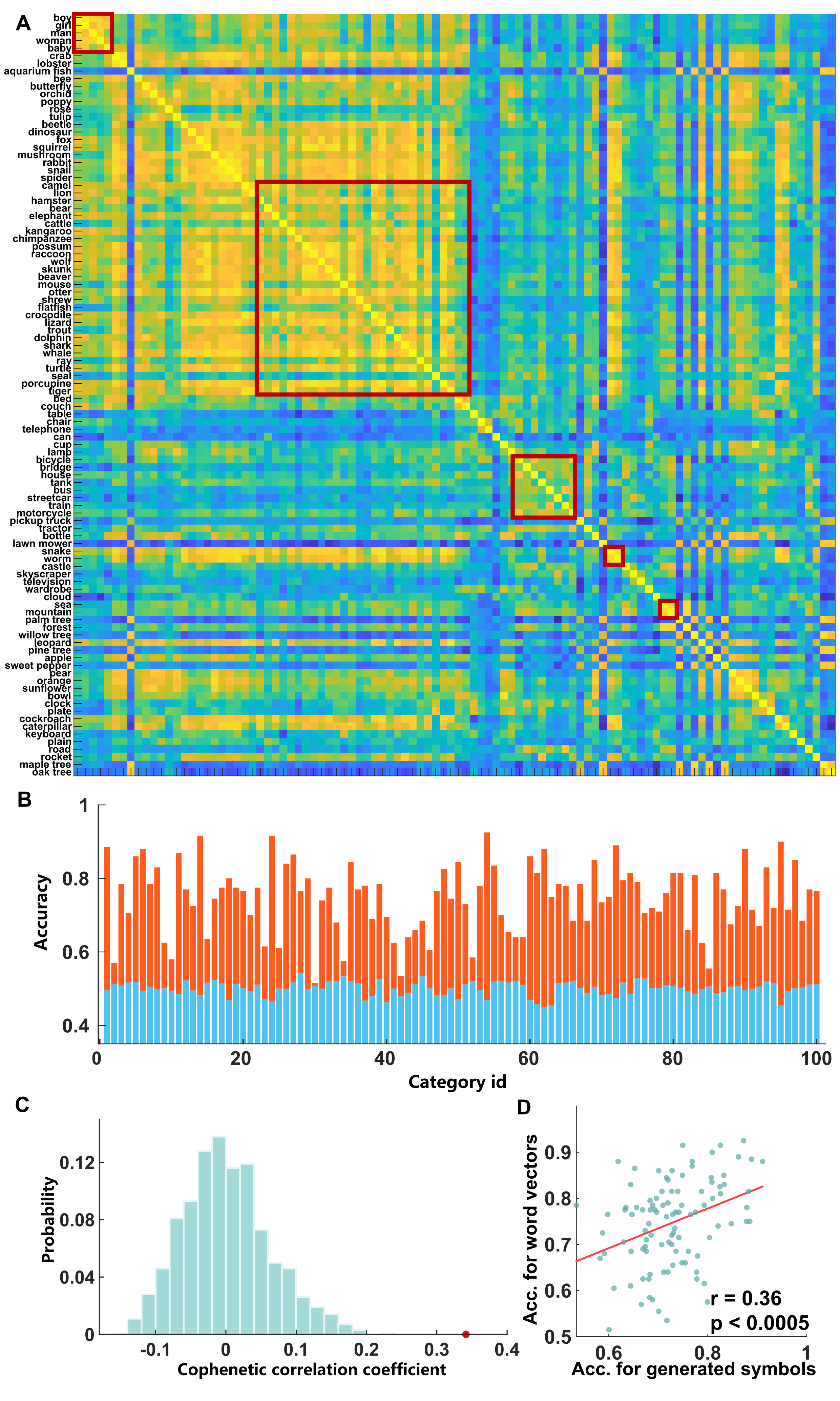}
	\captionsetup{labelformat=empty}
	\caption{}
\end{figure}
\begin{figure}
	\ContinuedFloat
	\caption{\textbf{Compatibility of SEA-net framework with natural language.}. 
	\textbf{(A)} Cosine similarity matrix among word vectors of category names. Order of categories is arranged according to clustering results of SEA-net-generated symbols shown in Fig. 2a. Red squares highlight several clusters also found in Fig. 2A. 
	\textbf{(B)} Identification accuracy (orange bar) of SEA-net on novel class images when informed by word vectors of category names. Light blue bar represents accuracy of random symbols. 
	\textbf{(C)} Distribution of co-phenetic correlation coefficients (see Methods for details) between shuffled symbols and word vectors (blue bars). Co-phenetic correlation coefficient between dendrogram of SEA-net-generated symbols and distances of word vectors is marked as a red dot. 
	\textbf{(D)} Scatter plot of accuracy of novel classes in (B) vs. that in Fig. 3C. Red line: linear regression.}
\end{figure}

\end{document}